%% file: preprint.tex
\documentclass{article}

\usepackage{arxiv}

\usepackage[utf8]{inputenc} 
\usepackage[T1]{fontenc}    
\usepackage{hyperref}       
\usepackage{url}            
\usepackage{booktabs}       
\usepackage{amsfonts}       
\usepackage{nicefrac}       
\usepackage{microtype}      
\usepackage{lipsum}
\usepackage[numbers]{natbib}

\usepackage[english]{babel}
\usepackage[utf8]{inputenc}
\usepackage[T1]{fontenc}

\usepackage{lineno}
\usepackage{hyperref}
\usepackage{amsmath}
\usepackage{graphicx,subcaption}
\usepackage{siunitx}
\usepackage{microtype}
\modulolinenumbers[5]
\usepackage{float}
\usepackage{etoolbox}

\title{Sub-pixel matching method for low-resolution thermal stereo images}

\author{
  Yannick Wend Kuni Zoetgnande\\
  Univ Rennes, Inserm,  LTSI - UMR 1099,\\
  F-35000 Rennes, France \\
  \texttt{yannick.zoetgnande@univ-rennes1.fr} \\
   \And
 Geoffroy Cormier \\
  Neotec Vision, \\
  35740 Pacé, France \\
  geoffroy.cormier@neotec-vision.com
  \And
 Alain-Jérôme Fougères \\
  ECAM Rennes - Louis de Broglie,\\
  35170 Bruz, France \\
  alain-jerome.fougeres@ecam-rennes.fr
  \And
 Jean-Louis Dillenseger \\
  Univ Rennes, Inserm,  LTSI - UMR 1099,\\
  F-35000 Rennes, France \\
  \texttt{jean-louis.dillenseger@univ-rennes1.fr} \\
}

\newbool{PREPRINT}
\booltrue{PREPRINT}

\begin{document}

\include{main_include}

\end{document}

%% file: main_include.tex
\ifbool{PREPRINT}{
    \maketitle
}{
  
}

\begin{abstract}
In the context of a localization and tracking application, we developed a stereo vision system based on cheap low-resolution $80 \times 60$ pixels thermal cameras. 
We proposed a threefold sub-pixel stereo matching framework (called ST for Subpixel Thermal): 1) robust features extraction method based on phase congruency, 2) rough matching of these features in pixel precision, and 3) refined matching in sub-pixel accuracy based on local phase coherence. We performed experiments on our very low-resolution thermal images (acquired using a stereo system we manufactured) as for high-resolution images from a benchmark dataset. Even if phase congruency computation time is high, it was able to extract two times more features than state-of-the-art methods such as ORB or SURF. We proposed a modified version of the phase correlation applied in the phase congruency feature space for sub-pixel matching. Using simulated stereo, we investigated how the phase congruency threshold and the sub-image size of sub-pixel matching can influence the accuracy. We then proved that given our stereo setup and the resolution of our images, being wrong of 1 pixel leads to a $\pm 500 $ mm error in the Z position of the point. Finally, we showed that our method could extract four times more matches than a baseline method ORB + OpenCV KNN matching on low-resolution images. Moreover, our matches were more robust. More precisely, when projecting points of a standing person, ST got a standard deviation of $\approx$ 300 mm when ORB + OpenCV KNN gave more than 1000 mm. 
\end{abstract}

\ifbool{PREPRINT}{
}{
\maketitle
}

\section{Introduction}

\label{sec:Introduction}

Detection of people is a crucial task in computer vision for security or safety applications (intrusion detection, pedestrian collision detection, fall detection). In this context, thermal cameras are often considered because they are robust to illumination changes, and they ensure the preservation of anonymity \cite{pittaluga2016sensor}. In some of these applications, the localization and tracking of people in the space is also a crucial issue. In this case, stereo-vision is often chosen because it produces accurate spatial information about the tracked people. Some works have already been performed on thermal stereo-vision based on high-resolution thermal cameras~\cite{Bertozzi2005, Dhua2003}, but one of the obstacles on the democratization of thermal cameras is their cost~\cite{priceIR}. Recently, a few manufacturers proposed very low-cost thermal cameras, e.g., the FLIR Lepton 2 \footnote{https://flir.netx.net/file/asset/19205/original}. The counterpart of these cameras is their low spatial resolution ($80 \times 60$ pixels for the Lepton 2). Such resolution has a direct impact on several steps of the traditional stereo-vision framework: stereo calibration, information extraction, information matching between the two views and triangulation. In ~\cite{Zoetgnande2018}, the authors proposed a robust low-resolution thermal stereo camera calibration. Matching is a prior step of 3D vision using a stereo system. The matching can be dense or sparse. Dense matching works well for textured images; unfortunately, thermal images are lacking texture. Thus, sparse matching on features extracted from the images must be considered. \textbf{But given a thermal image how to characterize the information it contains?}

A feature can be either global or local. Generally, global features are used in image retrieval, object detection, and classification. They are based on the fact that humans can easily recognize objects with a single glance~\cite{Olivia2006}. The most common global features descriptors are Histogram of Oriented Gradient~\cite{Bertozzi2007a,Olmeda2012}, Invariant moments~\cite{Zhang2009,Dai1999} and Co-occurrence Histograms of Oriented Gradients~\cite{Watanabe2009,Iwata2014}. The main drawback of these features is that their extraction is difficult and computationally costly. Unlike global features, local features are easier to extract. The most common local features methods are Harris corner detector~\cite{Harris}, KLT~\cite{Tomasi1993}, phase congruency~\cite{Morrone1987,Kovesi1999}, FAST~\cite{Rosten2008} and BRIEF~\cite{Calonder2010}. In the specific context of thermal images, the features can be sometimes merely extracted by thresholds and template matching~\cite{Liu2004}, using an improved version of the Harris corner detector~\cite{Dhua2003} or using the Canny edge detector~\cite{Bertozzi2007}. However, in~\cite{Hajebi2006}, the authors compare some features detectors such as Harris, Canny, Difference-of-Gaussian with the phase congruency. They prove that, regarding thermal images, phase congruency can extract more features than the methods as mentioned earlier. \textbf{ But after having extracted features, how to exploit them in stereo vision?}

The idea behind the stereo-vision is to exploit the difference between the two views, knowing more or less the position of each camera relative to the other one. Several method families are proposed in the literature for stereo computation: global methods such as dynamic programming~\cite{Ohta1985}, intrinsic curves~\cite{Brown2003} or local ones including block matching~\cite{saxena2019imaging}, gradient-based optimization~\cite{yang2009linear} and features matching~\cite{Baumberg2000}. Among these methods, features matching seems to be the most adapted to a sparse features situation. If most of the proposed methods have been developed for visible images, other authors propose an adaptation to thermal images. In~\cite{Dhua2003}, after having extracted Harris corners, the authors perform features matching by computing correlation within a search window, discarding outliers, and regularizing the matched features spatially. In~\cite{Bertozzi2005, Bertozzi2007}, among several variants, the authors propose a relatively similar framework for pedestrian detection: after having extracted features with the Canny edge detector, they perform matching using cross-correlation and specialized methods adapted to a human silhouette. Features matching is also important for thermal image analysis in medical applications \cite{dey2017thermal},  \cite{saxena2019study}. Most of the time, Thermal images are used to analyze the variation of the temperature of the skin or an organ. So in such a condition, the extracted features must be very robust. In \cite{saxena2019infrared}, the authors propose a framework to correct motion artifact due to body movement when recovering breast images. While the amplitude of the Fourier transforms gives the pixel intensity, the phase components represent the spatial information. Given three consecutive frames $I_1$, $I_3$ and $I_3$, they combine their phase and amplitude to get a final matched image $I'_3$. The matching is performed using the method described in \cite{reddy1996fft}. In~\cite{Hajebi2006}, the authors, extract the features from thermal images using Log-Gabor filters bank and then perform 1D matching through epipolar lines using the Lades similarity~\cite{Lades1993} as a matching criterion. So, processing thermal images using Fourier-based methods is a common way. 

All of these works concern thermal images with a reasonable pixel resolution (over $80 \times 60$). In~\cite{Hajebi2008}, the authors state that the computation time of the phase congruency can be reduced by down-scaling the original from $320 \times 240$ to $160 \times 120$ and $80 \times 60$, but they have not explored this solution given the induced loss of efficiency. In our specific case of stereovision, the image resolution has a direct impact on the 3D reconstruction. Indeed being wrong of 1 pixel in a $4  \times 4$ image is not the same as being wrong of 1 pixel in a $1000 \times 1000$ image. The information matching between left and right views of the stereo pair is a critical phase, since small errors at this step may yield significant errors in the 3D localization as demonstrated below.  The distance between a 3D point and the stereo system can be determined as follows \cite{delon2007small}:

\begin{equation}
\label{eq:z}
    z = \frac{\epsilon}{b/h}
\end{equation}

where $b$ is the baseline of the stereo system (distance between the two cameras), $\epsilon$ the disparity function and $h$ the distance between the scene and the camera system. Deriving \ref{eq:z}, we have: 

\begin{equation}
    dz = \frac{d\epsilon}{b/h}
\end{equation}

The error $d \epsilon$ on the disparity has a direct impact on the precision in z, especially when $b/h$ is small. The influence of the baseline on the depth precision has been well studied in the literature. In \cite{maier2013optical}, the authors empirically show that to determine the distance of a point accurately, the latter must not be further away more than ($10-15$) $\times$ baseline. Moreover, in \cite{haller2011design}, the authors stated that using their stereo setup, a disparity error larger than 0.1 pixels will result in a relative distance error of 2.5\% for an object located at 60 meters. This is why, most of the time, the stereo-vision system uses a large baseline \cite{pritchett1998wide}. In this application, we chose to use a small baseline given indoor constraints. Nevertheless, some applications (as our indoor system) are constrained to use a small baseline. \textbf{Is it possible to gain in  3D accuracy even in case of small baseline and low image resolution?}

Many solutions exist to improve the accuracy of stereo-vision. Among them we have super-resolution \cite{zoetgnande2019edge} and sub-pixel matching \cite{tian1986algorithms}. Considering sub-pixel matching, we must first define the disparity map, which \textit{"refers to the apparent pixel difference or motion between a pair of a stereo image"} \cite{smith2018processing}.  In the classical methods, the difference is expressed in integer precision. Sub-pixel matching proposes to estimate e in floating precision. There are three main methods of sub-pixel matching: interpolation, fitting, and phase correlation. In interpolation methods, the sub-pixel disparity is estimated by searching the extreme of interpolated cost volume. These methods have two main drawbacks. First, the interpolation of the cost volume is computationally costly. Second, artifacts can be introduced by the interpolation even if most of the time, the accuracy is satisfactory~\cite{haller2012design,miclea2015new}. The fitting methods use disparity plane or cost volume to estimate the sub-pixel disparity. The methods using disparity plane fitting involve segmentation constraints~\cite{Shi2015}. The cost volume fitting methods search the extreme of a parabola representing this cost volume. These methods are fast but not accurate. Unlike these two methods, phase correlation offers both high efficiency and accuracy. It is performed using fast Fourier transforms and other supplementary approaches~\cite{Stone2001}. The phase correlation is the normalized cross-power spectrum, so the matching corresponds to a peak, which has to be estimated~\cite{Takita2003} in sub-pixel accuracy. In the state of the art, all of the methods are performed for visible images. We proposed to compute the sub-pixel disparity for thermal images using phase correlation. To our knowledge, it is the first time a sub-pixel matching is performed in such a way in thermal images.

The paper is structured as follows: Section~\ref{sec:SubPixelMatchingFramework} presents our framework ST. It details the implementation of the features extraction method for thermal images based on phase congruency, the stereo matching method, and the sub-pixel matching method based on phase correlation. In Section~\ref{sec:Discussion}, we discuss, analyze, and explain these results. Finally, section~\ref{sec:Conclusion} concludes the paper and gives perspectives.


\section{Materials and method}
\label{sec:SubPixelMatchingFramework}

\subsection{Our stereo system}

\begin{figure}[]
\centering
  \includegraphics[width=\linewidth]{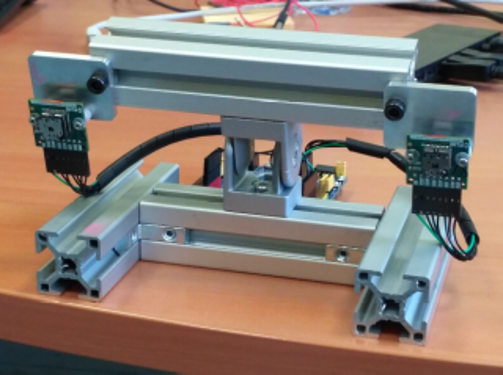}
  \caption{The stereo system composed of two lepton 2 cameras placed.}
  \label{fig:stereo_cameras}
\end{figure}

Our indoor acquisition system is composed of a pair of FLIR's lepton 2 cameras (Fig. \ref{fig:stereo_cameras}). The horizontal field of view is $51^{\circ}$ and the diagonal field of view is $63.5^{\circ}$. The maximum frame rate is eight frames per second. The baseline is set to 16 mm.

As input images, we had a pair of low-resolution thermal images acquired in stereo condition $I_l$ and $I_r$. The images are rectified, using the output of a robust calibration method~\cite{Zoetgnande2018}. The rectification step simplifies the stereo reconstruction step since that a feature lies now on the same line in $I_l$ and $I_r$. The feature matching (and so the estimation of the disparity) is reduced to only estimate the translation $\delta_x$ of the feature  between the 2 images along the x direction. We have: $\delta_x = d_i(x) + d_d(x)$ where $d_i(x)$ and $d_d(x)$ are respectively the integer and the decimal parts of the disparity.

\begin{figure*}
\centering
  \includegraphics[width=1\linewidth]{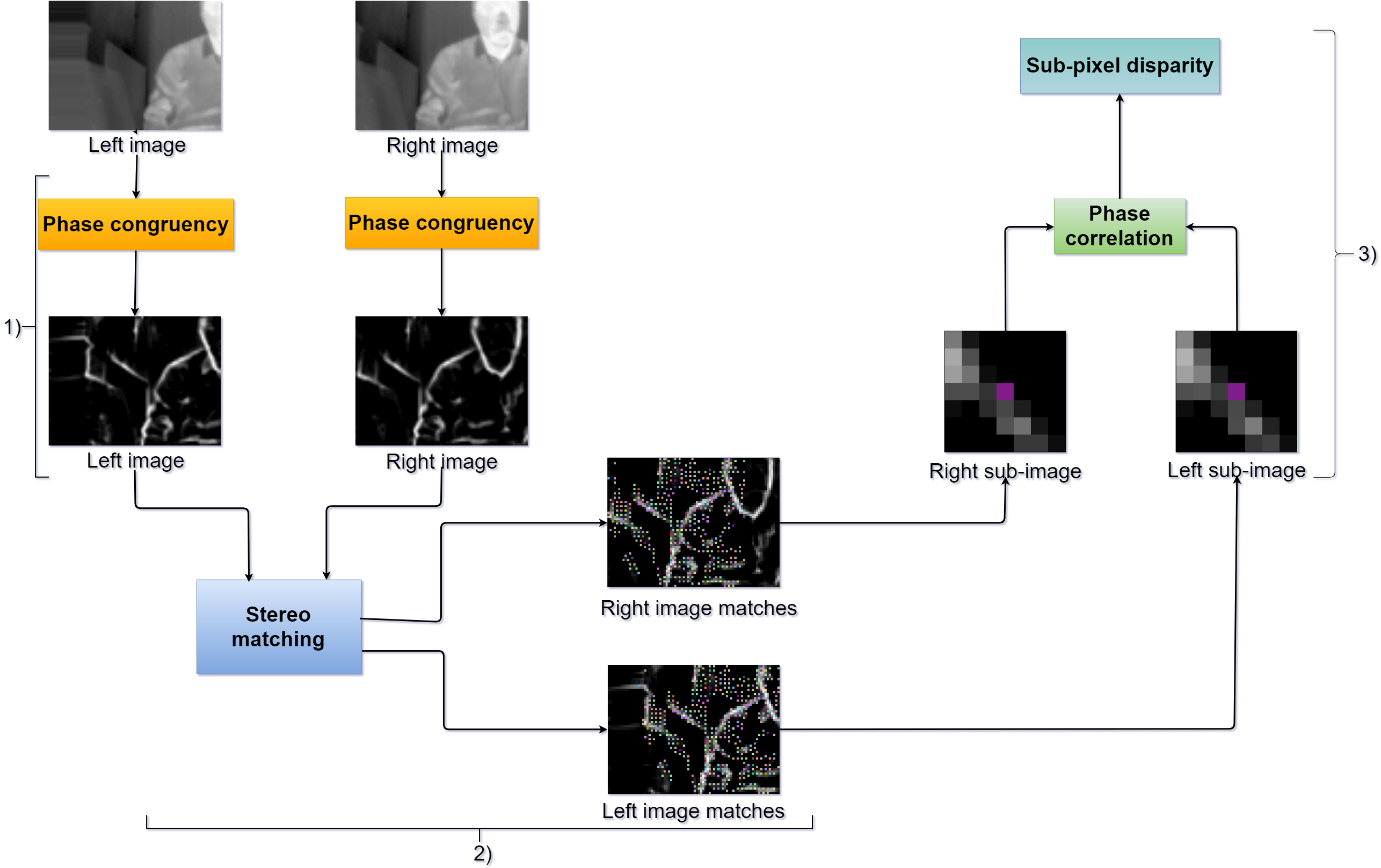}
  \caption{Sub-pixel matching framework: 1) robust features extraction; 2) rough features matching; 3) refined matching in sub-pixel accuracy.}
  \label{fig:subPixelMatching}
\end{figure*}

Our sub-pixel matching framework is composed of three main steps (Fig.~\ref{fig:subPixelMatching}): 1) robust features extractions method from the low-resolution thermal images, 2) first rough stereo matching of these features in integer precision (the estimation of $d_i(x)$) and 3) refined sub-pixel matching around the previously matched features (the estimation of $d_d(x)$). 

\subsection{Features extraction}
\label{ssec:FeaturesExtractionMeth}

Features extraction was one of the critical points of our framework. Thermal images are characterized by a lack of texture, noisy, and since the resolution of our images is low, a robust method is necessary to extract features from such images. Besides the low-resolution aspect, other characteristics of our images had to be taken into account. Our cameras are uncooled, and thus, they are influenced by ambient temperature. The temperature drift is sometimes compensated in the camera by some corrections which introduce sudden brightness changes. More, this is camera dependent and is not synchronized, which leads to a time-varying difference in brightness between the images. For all these reasons, we chose a method based on phase congruency, which takes into account primitives that are only linked to the difference of viewpoint between the two cameras.

To extract features, we adapted the phase congruency estimation method proposed by~\cite{Kovesi2000} and its variant proposed by ~\cite{Wang2014} to our low-resolution thermal images.
 The 1D phase congruency is the ratio~\cite{Morrone1987}:

\begin{equation} \label{eq:phaseCong}
  PhaseCong(x) = \max_{\bar{\phi}} \frac{\sum_{n} A_{ n} \cos(\phi_{ n}(x)-\bar\phi)}{\sum_{ n}A_n}
\end{equation}

Where $A_n$ represents the amplitude or energy of the $n^{th}$ Fourier component and $\phi_{ n}(x)$ the local phase, which can be calculated using the Hilbert transform. Unfortunately, this version of the phase congruency was noise sensitive and yielded inaccurate features localization as proved by \cite{kovesi1999image}. In~\cite{kovesi2003}, the authors modified a bit the equation to circumvent  the previous version drawbacks:

\begin{equation} \label{eq:phaseCong2}
\begin{split}
PhaseCong_2(x) = W(x) \frac{ \left \lfloor \sum_{n} A_n \left [ \Upsilon_n - \Psi_n \right ] - T \right \rfloor}{ \sum_n A_n(x) - \epsilon } 
\end{split}
\end{equation}

where $\Upsilon_n = \left | \cos(\phi_n(x) - \bar{\phi(x)}) \right | $; $\Psi_n = \sin(\phi_n(x) - \bar{\phi(x)}) $, $W(x)$ is the frequency spread weighting, $T$ is a noise threshold and $\epsilon$ is a small value (eg. 1-e3) to avoid division by 0.

The equation (\ref{eq:phaseCong2}) can be extended to the two-dimensional image domain by applying it on several orientations $\theta$ after filtering the image by a bank of Log-Gabor filters. To reduce the computation cost due to the number of orientations and scales of the Log-Gabor filters bank, we implemented the variant proposed by~\cite{Wang2014} using a monogenic filter instead of Log-Gabor. The monogenic signal is a Riesz transform concatenated with a 2D signal. This was possible by constructing a monogenic filter in the frequency domain \cite{Wang2014}. 

As results of the 2D extension, we get a set of $PC(\theta)$, the phase congruency at orientation $\theta$. The features are then estimated by combining all the $PC(\theta)$. This is done by computing the following values~\cite{kovesi2003}:

\begin{align}
a = \sum (PC(\theta) cos(\theta))^2\\
b = 2 \sum (PC(\theta) cos(\theta)) ( PC(\theta) sin(\theta) )\\
c = \sum (PC(\theta) sin(\theta))^2
\end{align}

Combining $a$, $b$ and $c$ gives a hint of the strength of the feature. More precisely, the maximum moment $M$ and the minimum moment $m$ can be estimated by:

\begin{align}
 \label{eq:magnitudeM}
M = \frac{1}{2} ( c + a + \sqrt[2]{b^2 + (a -c)^2} ) \\
\label{eq:magnitudem}
m = \frac{1}{2} ( c + a - \sqrt[2]{b^2 + (a - c)^2} )
\end{align}

These moments are used to characterize the features. Higher is the maximum moment more significant will be the feature, and higher is the minimum moment more probably this feature point will be a corner. Because of the lack of information in images, we only took $M$ into accounts. So a feature was considered significant if $M$ was higher than a threshold $\gamma$. 

As a result of this step, we produced two images $I_{fl}$ and $I_{fr}$, which are the images of the moment $M$ after applying the phase congruency on respectively the input images $I_l$ and $I_r$ (Fig.~\ref{fig:subPixelMatching}).

\subsection{Stereo matching}
\label{ssec:StereoMatchingMeth}

In rectified image condition, the matching of a feature $F^{l}$ of $I_{fl}$ to its corresponding one $F^{r}$ in $I_{fr}$ is simplified to find the most similar feature along the corresponding epipolar line. As suggested by~\cite{Hajebi2006}, we used the Lades similarity~\cite{Lades1993} performed a $5 \times 5$ window as matching metric: 

\begin{equation}\label{eq:lades}
S(F^{l},F^{r}) = \frac{ \sum_j^{nb_{feat}} f_j^l f_j^r }{ \sqrt[2]{ \sum_j^{nb_{feat}} {f^l}_j^2 \sum_j^{nb_{feat}} {f^r}_j^2 } }
\end{equation}

where $nb_{feat}$ is the total number of features in the selected window.

However, compared to their work, we took into account more matching constraints than only the similarity, epipolar constraints and left-rigth consistency:
\begin{itemize}
\item Uniqueness: a feature in the left image was matched with only one feature in the right image.
\item Continuity: the disparity must vary smoothly. 
\item Ordering: for a couple of matched features $f_{1l} \leftrightarrow f_{1r}$ , $f_{2l} \leftrightarrow f_{2r}$ (the symbol $\leftrightarrow$ represents the matching relationship), if $f_{1l}$ is at the left of $f_{2l}$ we ensured that $f_{1r}$ was at the is of $f_{2r}$. 
\end{itemize}

We also took into account the type of features and their orientation in the matching process.

We tried to reduce the computation time using a disparity range of $d$. This range can be obtained if there is prior knowledge about the scene or it can be deduced by the disparities values of the previous frame in a frame by frame framework.

\subsection{Sub-pixel matching}
\label{ssec:SubPixelMatchingMeth}

Phase correlation is a well-known method allowing to get the translation of a signal. The translation between two images in the spatial domain is expressed in the frequency domain using the Fourier transform. Let $I_1$ and $I_2$ be two $M \times N$ images, $M$ and $N$ odd for mathematical simplicity. These images can be expressed like $I_1 = I(x,y)$ and $I_2 = I(x + \delta x, y + \delta y)$ with the translation $(\delta x, \delta y)$. Let $m$ and $n$ defined as $M = 2m+1$ and $N = 2n+1$. By computing their respective Fourier transform $F_1(u,v)$ and $F_2(u,v)$, the normalized cross-power spectrum is given by:

\begin{equation} \label{eq:normalizedCrossPowerSpectrum}
R(u,v) = \frac{ F_1(u,v) F^{\star}_2 (u,v) }{ \left | F_1(u,v) F^{\star}_2 (u,v) \right | } 
\end{equation}

where $u = -m,...,m$, $v = -n,...,n$ and $F^{\star}_2$ is the complex conjugate of $F_2$.

Given that the most relevant components in the phase correlation matrix are the low-frequency ones, certain authors propose to filter $R$ by a rectangular low-pass function of size $U$~\cite{Takita2003}. They also proved that the ratio of $\frac{U}{M} = 0.5$ is the one that gives the best accuracy. 

The Phase-only correlation is the inverse discrete Fourier transform of $R$ and is defined as follows: 

\begin{equation} 
\begin{split}
    \label{eq:poc}
r(x,y) = \frac{1}{N \cdot M} \sum_{u = -m}^{u = m} \sum_{v = -n}^{v = n} R(u,v) \\ \exp(- 2 \pi i u x /M) \exp(- 2 \pi i v y /N) 
\end{split}
\end{equation}

where $u = -m,...,m$ and $v = -n,...,n$.

The peak position, in $r$, corresponds to the translation along $x$ and $y$ directions, but this position is in integer precision. 

Some authors propose to recover the displacement in sub-pixel precision from the information contained in $r$. There are globally three ways to estimate the sub-pixel displacement accurately: the detection of the peak by local least square fitting on selected spectral components in $R$~\cite{Stone2001}, by fitting a closed-form analytic model to the correlation peak~\cite{Takita2003}, or directly from the data of $r$ using a peak detection formula assuming a peak model~\cite{Foroosh2002,Nagashima2006}. We implemented this last principle.

The correlation peak was modeled as a cardinal sine affected by the low pass filter:
\begin{equation}
\label{eq:peakmodel2D}
\begin{split}
       PeakModel_{xy} = r(x,y) \approx \alpha \frac{ \sin(\frac{V}{M}\pi (x + \delta_x) ) }{ \pi (x + \delta_x) } \\ \times \frac{ \sin( \frac{V}{N} \pi (y + \delta_y) ) }{ \pi (y + \delta_y) }
\end{split}
\end{equation}

where $\delta_x$,$\delta_y$ are respectively the displacements along $x$ and $y$ axis, $\alpha$ is the peak value ( $0 \leq \alpha \leq 1$) and $V = 2U +1$.

Given that we used stereo cameras, by rectifying the images, we were able to neglect $\delta_y$. So the phase correlation function became: 

\begin{equation}\label{eq:peakmodel1D}
PeakModel_x =    r(x) \simeq \alpha \frac{\sin (\frac{V}{M}\pi (x + \delta_x))}{\pi (x + \delta_x)}
\end{equation}

\begin{figure*}
    \centering
        \includegraphics[width=1\textwidth]{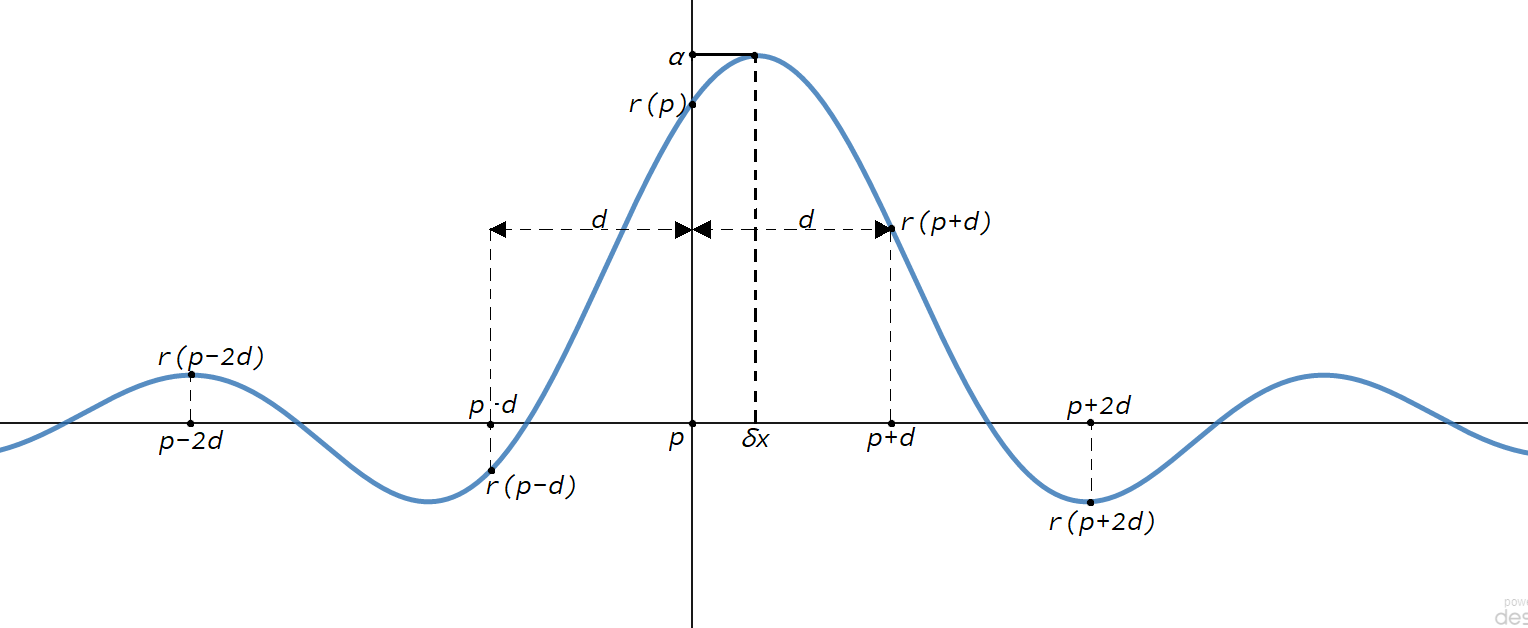}
    \caption{Peak model. In the vertical axis, the value of the phase correlation value at the point in the horizontal axis.}
    \label{fig:good_pp1}
\end{figure*}

The main idea was then to recover $\delta_x$, the max peak in sub-pixel accuracy, from the sample values in $r$. A robust solution has been described in~\cite{Nagashima2006}. Let consider the case in Fig.~\ref{fig:good_pp1}, the peak model max is close to the sample $x = p$ which has the highest values $\alpha$. We also considered two other measurements located at $\pm d$ pixels away from $p$ at $x = p + d$ and $x = p - d$.

Using the model, the three points can be rewritten as:

\begin{equation}\label{eq:peakmodelNeighbors}
    \begin{cases} 
        PeakModel_x(p-d) = r(p-d) \simeq \alpha \frac{\sin (\frac{V}{M}\pi (p -d + \delta_x))}{\pi (p -d + \delta_x)}\\
        PeakModel_x(p) = r(p) \simeq \alpha \frac{\sin (\frac{V}{M}\pi (p + \delta_x))}{\pi (p + \delta_x)} \\ 
        PeakModel_x(p+d) = r(p+d) \simeq \alpha \frac{\sin (\frac{V}{M}\pi (p+d + \delta_x))}{\pi (p+d + \delta_x)}
    \end{cases}
\end{equation}

By combining these equations, it can be proved that:

\begin{equation*}
\begin{split}
        \left ( p - d + \delta_x \right ) r\left ( p - d \right ) + \left ( p + d + \delta_x \right ) r\left ( p + d \right ) \\ - 2 \left ( p + \delta_x \right ) \cos \left ( \frac{V}{N} \pi d \right ) r \left ( p \right ) = 0
\end{split}
\end{equation*}

which can be rewritten as 

\begin{equation} \label{eq:peakmodelNeighborsCombined}
    v(p,d) = \delta_x u(p,d)
\end{equation}

with

\begin{equation*}\label{eq:peakmodelNeighborsCombinedTerms1}
\begin{split}
        u(p, d) = r(p - d) + r(p + d) - 2 \cos\left ( \frac{V}{N} \pi d \right ) r\left ( p \right )
\end{split}
\end{equation*}

and 

\begin{equation*}\label{eq:peakmodelNeighborsCombinedTerms2}
\begin{split}
v\left ( p, d \right ) = 2p \cos\left ( \frac{V}{N} \pi d \right ) r\left ( p \right ) \\- \left ( p - d \right ) r\left ( p - d \right ) - \left ( p + d \right ) r\left ( p + d \right )
\end{split}
\end{equation*}

This allows us to estimate $\delta_x = u\left ( p,d \right )^{-1} v\left ( p,d \right )$.

However, in order to minimize the impact of noise, several observations with different values of $p$ and $d$ around the highest peak in r can be used. So it is possible to select $\chi$ values of $p_i$ and $d_i$ and then get $\chi$ equations $v\left ( p_i, d_i \right ) = \delta_x u\left ( p_i, d_i \right )$ where $i \in \left \{1, 2, ..., \chi \right \}$. Resolving these equations is equivalent to minimize $\delta_x$ in:

\begin{equation} \label{eq:minSquareDeltaX}
 \delta_x = \sum_{i = 1}^{\chi} \left | v\left ( p_i, d_i \right ) - u\left ( p_i, d_i \right ) \right |^2
\end{equation}

This equation can be solved using Singular Value Decomposition (SVD) \cite{abdi2007singular}.

The existing phase correlation-based sub-pixel matching methods are applied on high resolution visible images (around $3000 \times 3000$ pixels~\cite{Ma2017}) using large sub-windows ($41 \times41$ pixels) for the phase correlation estimation. We had to adapt this class of methods to our low-resolution problem. Let consider two matched points $F_l$ with coordinate $(x_l,y_l)$ and $F_r$ with coordinate $(x_r,y_r)$ respectively in $I_{fl}$ and in the shifted image $I_{fr}$. We knew that $y_l = y_r$ and $d_i(x) = x_r - x_l$. Then we defined two sub-images $Is_l(x_l,y_l)$ and $Is_r(x_r,y_r)$ with the same windows size $W$ around $F_l$ and $F_r$ (Fig.~\ref{fig:subPixelMatching}) and performed the refined phase correlation-based matching in these sub-images. 

The main parameters of this method are the sub-images windows size $W$ and the number $\chi$ of observations used for the least square estimation of the sub-pixel displacement (\ref{eq:minSquareDeltaX}). Because of our low resolution we limited the number of observations $\chi$ to 6: $p_i = p \pm 1$ and $d_i \in \left\{1,2 \right\}$. 

 \section{Results and discussion}
\label{sec:Discussion}

\subsection{Datasets}

In this paper, we used two datasets: a thermal infrared video benchmark for visual analysis with high-resolution infrared images \cite{Wu2014} and an own dataset (called Tvvlgo in the rest of the paper) with images acquired by our system. Tvvlgo was created by placing the stereo system in the ceil of a room, and we collected 1000, 80x60 pixels images pair of a person moving in a room. We also used another third image where a person is sitting in from of the cameras. The dataset we used is available online \cite{VideoTraz}.

\subsection{Feature extraction}
\label{ssec:ExpFeatureExtraction}

In~\cite{Kovesi1999}, the author proposes to use a threshold on the phase congruency moments of $\gamma = 0.3$. Because of our low-resolution images context, we feared that this threshold could not extract enough features. We tried other smaller thresholds at $\gamma = 0.1$ and $\gamma = 0.01$. 

\subsubsection{Evaluation of the number of extracted features}

\begin{figure*}
     \centering
     \begin{subfigure}[b]{0.3\textwidth}
        \centering
        \includegraphics[width=\textwidth]{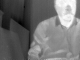}
        \caption[Network2]%
        {{\small Original image}}   
        \label{fig:original_8060}
     \end{subfigure}
     \hfill
     \begin{subfigure}[b]{0.3\textwidth} 
        \centering 
        \includegraphics[width=\textwidth]{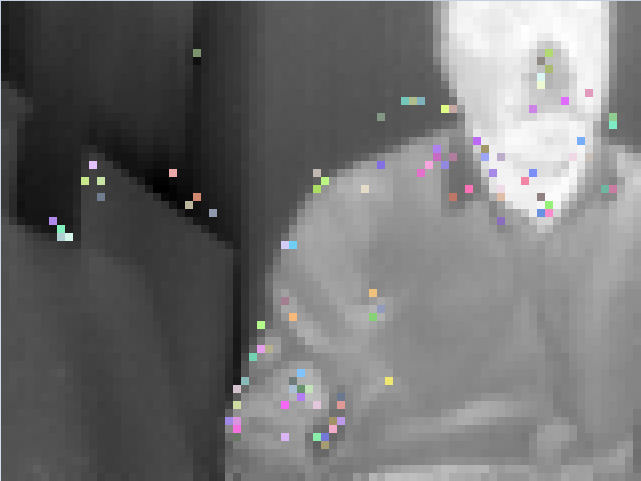}
        \caption[]%
        {{\small ORB }}   
        \label{fig:feature_orb}
     \end{subfigure}
     \hfill
     \begin{subfigure}[b]{0.3\textwidth} 
        \centering 
        \includegraphics[width=\textwidth]{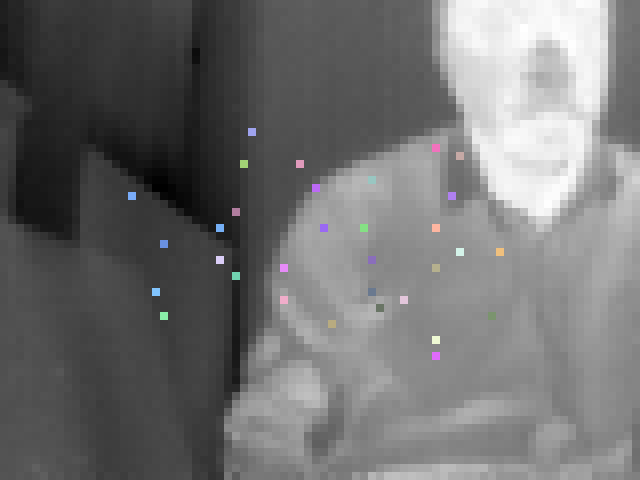}
        \caption[]%
        {{\small BRISK }}   
        \label{fig:feature_brisk}
     \end{subfigure}
     \vskip\baselineskip
		\begin{subfigure}[b]{0.3\textwidth}
        \centering
        \includegraphics[width=\textwidth]{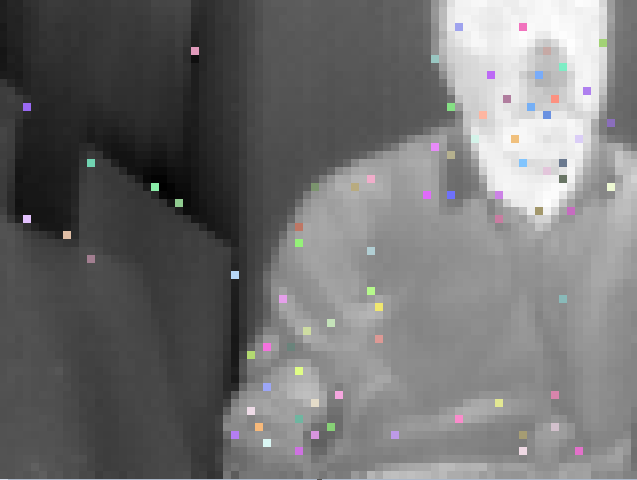}
        \caption[Network2]%
        {{\small FAST}}   
        \label{fig:feature_fast}
     \end{subfigure}
     \hfill
     \begin{subfigure}[b]{0.3\textwidth} 
        \centering 
        \includegraphics[width=\textwidth]{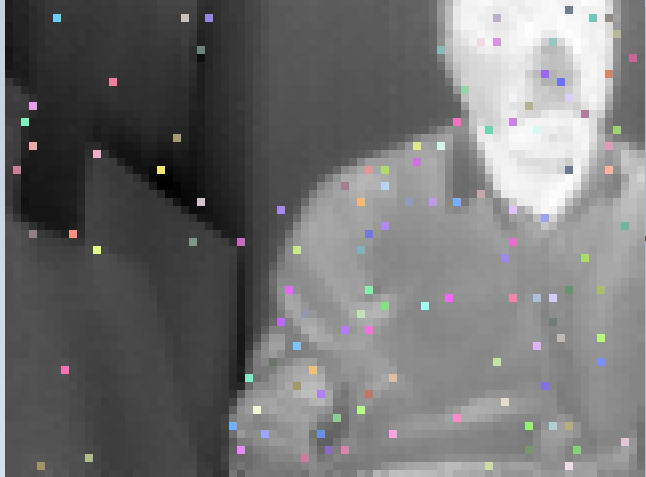}
        \caption[]%
        {{\small Shi and Tomasi }}   
        \label{fig:feature_shit_tomasi}
     \end{subfigure}
     \hfill
     \begin{subfigure}[b]{0.3\textwidth} 
        \centering 
        \includegraphics[width=\textwidth]{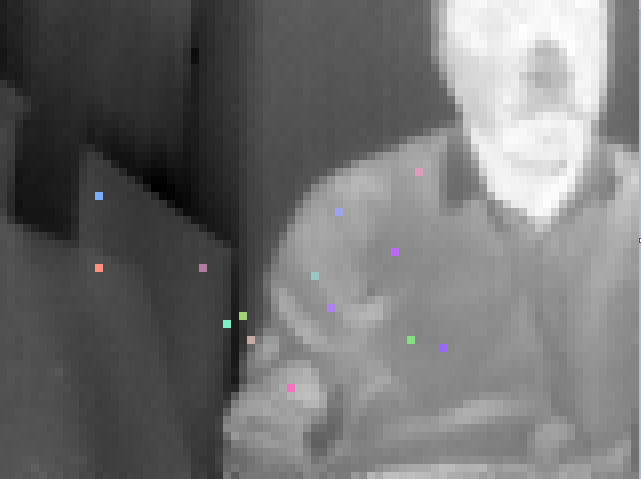}
        \caption[]%
        {{\small SURF }}   
        \label{fig:feature_surf}
     \end{subfigure}
     \vskip\baselineskip
     \begin{subfigure}[b]{0.3\textwidth}
        \centering
        \includegraphics[width=\textwidth]{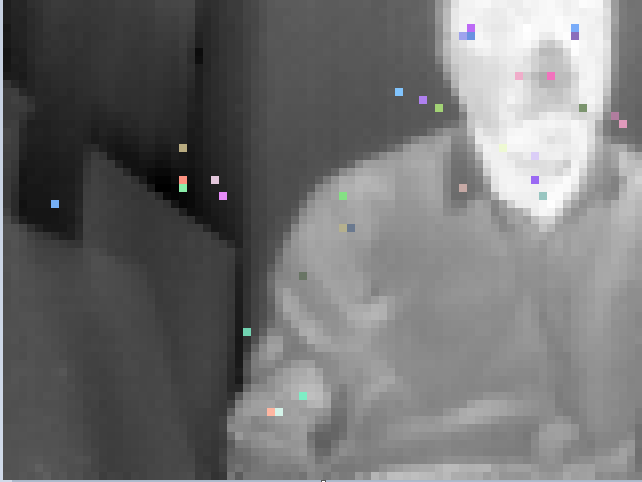}
        \caption[Network2]%
        {{\small KAZE }}   
        \label{fig:feature_kaze}
     \end{subfigure}
     \hfill
     \begin{subfigure}[b]{0.3\textwidth} 
        \centering 
        \includegraphics[width=\textwidth]{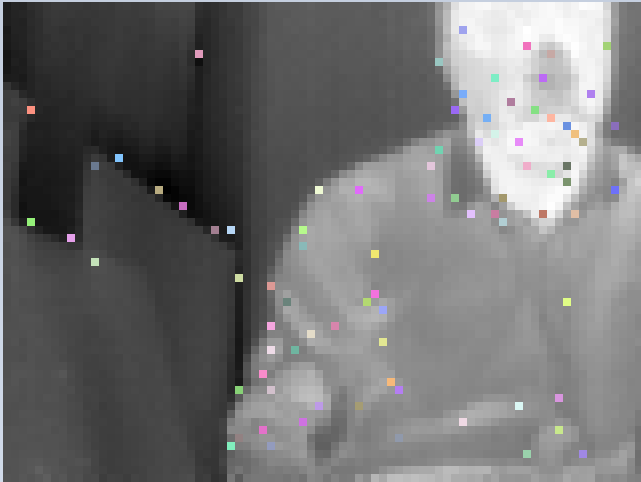}
        \caption[]%
        {{\small AGAST}}   
        \label{fig:feature_agast}
     \end{subfigure}
     \hfill
     \begin{subfigure}[b]{0.3\textwidth} 
        \centering 
        \includegraphics[width=\textwidth]{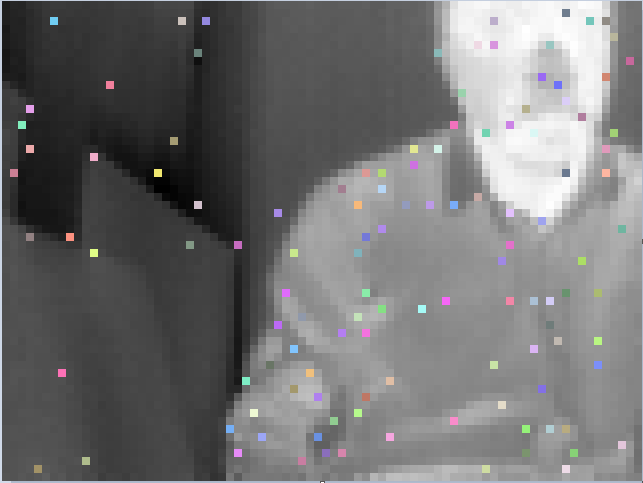}
        \caption[]%
        {{\small GFTT}}   
        \label{fig:feature_gftt}
     \end{subfigure}
     \vskip\baselineskip
           \begin{subfigure}[b]{0.3\textwidth}
        \centering
        \includegraphics[width=\textwidth]{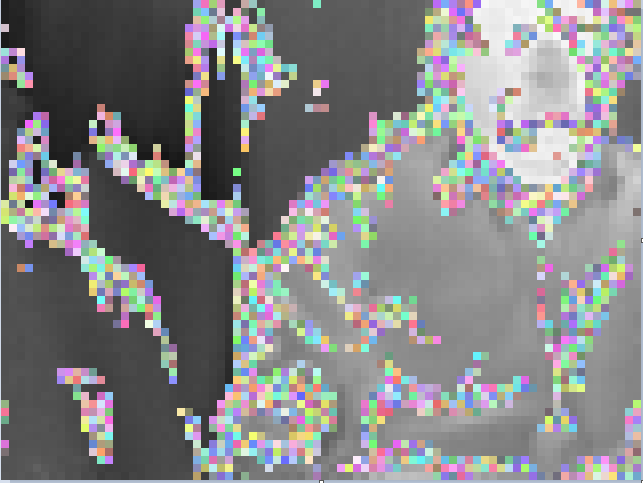}
        \caption[Network2]%
        {{\small PhaseCong(0.01) }}   
        \label{fig:feature_pc_001}
     \end{subfigure}
     \hfill
     \begin{subfigure}[b]{0.3\textwidth} 
        \centering 
        \includegraphics[width=\textwidth]{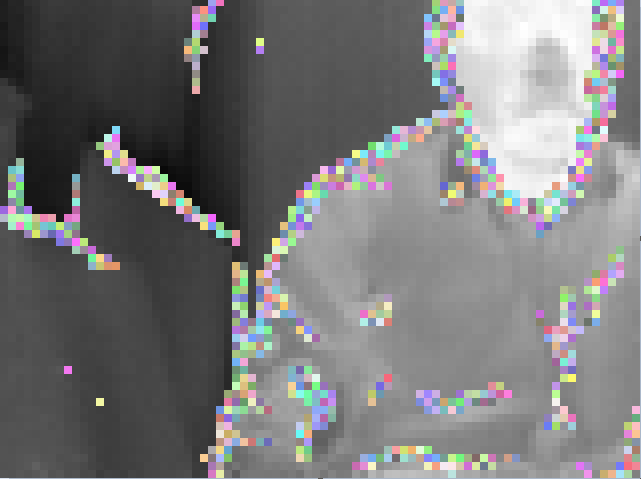}
        \caption[]%
        {{\small PhaseCong(0.1) }}   
        \label{fig:feature_pc_01}
     \end{subfigure}
     \hfill
     \begin{subfigure}[b]{0.3\textwidth} 
        \centering 
        \includegraphics[width=\textwidth]{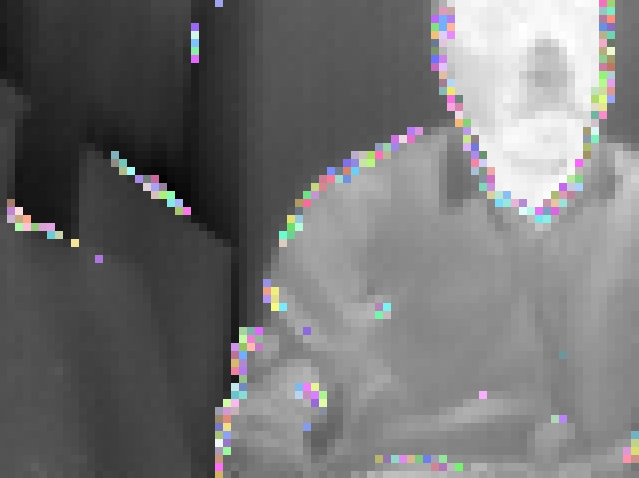}
        \caption[]%
        {{\small PhaseCong(0.3) }}   
        \label{fig:feature_pc_03}
     \end{subfigure}
     \caption[ The average and standard deviation of critical parameters ]
     {\small Features (colored dots) extracted by several feature extractor methods. The color of feature is generated randomly.} 
     \label{fig:feature_extraction_results}
   \end{figure*}

To validate our choice of the features extraction method, we compared the implemented phase congruency to other standard features extraction methods using a low-resolution image given by a FLIR lepton 2 (Fig.~\ref{fig:feature_extraction_results}-a). The authors in~\cite{Hajebi2008} have already made such comparison, but they only compared the phase congruency method to Harris corner detector, SIFT, Canny, and KLT. We wanted to go further. So we have compared our approach with the three thresholds (PhaseCong(0.01), PhaseCong(0.1) and PhaseCong(0.3)) to the OpenCV \cite{kaehler2016learning} implementations of ORB, BRISK, FAST, Shi Tomasi, SURF, AGAST, GFTT, and KAZE. Our framework was implemented using C++ compiled with GCC 7. We compared the methods according to 2 criteria: the number of extracted features and the execution time. For this later measurement, all the codes were executed on an Intel Core i7-3687U CPU. To get the execution time we used the C++ API \textbf{std::chrono::high\_resolution\_clock::now}. For each feature detector, the feature extraction process was completed \num{1000} times, and we computed the mean and the standard deviation of all computation times. 

The extracted features can be seen in Fig.~\ref{fig:feature_extraction_results} and measurements are sampled in Table~\ref{tab:compFeatures}. 

As illustrated by Table~\ref{tab:compFeatures} Phase congruency can extract more features than other methods. Even using a high threshold, we could extract two times more features than Shi-Tomasi and GFTT. The counterpart of this advantage is a relatively high computation time compared to techniques such as FAST. Fortunately, the feature extraction for a pair of images took approximately 10 ms with Phase congruency, which is compatible with the 8 frames per second rate of our cameras. 

These results are visually confirmed by Figure \ref{fig:feature_extraction_results}, where we extracted features from a low-resolution image using different feature extractors. It is noticeable that while most of the feature extractors try to extract robust features, phase congruency does the same by focusing on edges. Using the threshold $\gamma$, we could control the number of features returned by phase congruency.  $\gamma = 0.1$ seemed to give a good trade-off between the sparsity and redundancy of features.

To conclude, even if Phase congruency is slower to compute than the other classical method, it gave a higher number of features and so seemed to be unavoidable in processing our low resolution
and texture-less images.

\begin{table*}[]
\centering
\begin{tabular}{|l|l|l|}
\hline
Feature extractions methods                                                                                                    & Number of features & Execution time (us) \\ \hline
ORB (Oriented fast and Rotated Brief)                                                                                          & 117                & $ 245 \pm 20$                 \\ \hline
BRISK (BinaryRobust Invariant Scalable Keypoints)                                                                              & 34                 &    $ 373474 \pm 350$             \\ \hline
FAST                                                                                                                           & 77                 &   $ 90 \pm 7$                \\ \hline
Shi Tomasi                                                                                                                     & 120                &    $ 189 \pm 15$              \\ \hline
SURF (min Hessian = 300) (Speeded Up Robust Features)                                                                          & 14                 &   $ 1436 \pm 100$              \\ \hline
\begin{tabular}[c]{@{}l@{}}AGAST (Adaptive and generic corner detection \\ based on the accelerated segment test)\end{tabular} & 84                 &   $ 202 \pm 29$               \\ \hline
GFTT (Good Features to track)                                                                                                  & 120                &    $ 350 \pm 32$              \\ \hline
KAZE                                                                                                                           & 33                 &  $ 13217 \pm 123$              \\ \hline
PhaseCong(0.01)                                                                                                                & 1734               &   $ 5224 \pm 480$              \\ \hline
PhaseCong(0.01)                                                                                                                & 777                &   $ 5224 \pm 480$              \\ \hline
PhaseCong(0.01)                                                                                                                & 275                &   $5224 \pm 480$              \\ \hline
\end{tabular}
\caption{Comparison between feature extractor methods ORB, BRISK, FAST, Shi Tomasi, SURF, AGAST, GFTT, KAZE and Phase congruency}
   \label{tab:compFeatures}
\end{table*}

\subsubsection{Robustness to illumination change}

One particularity of our cameras is that we sometimes noticed a sudden change in the brightness. It is probably a re-calibration of the sensor. The feature extraction method should so be robust to brightness changes. We evaluated this robustness using a simulated brightness change as defined by Szelinski  \cite[Chapter~3]{szeliski2010computer}:

\begin{equation}
\label{eq:color}
    g(x) = \alpha f(x) + \beta
\end{equation}

where f(x) and g(x) are the pixel values, and $\alpha$, $\beta$ control respectively the contrast and the brightness.

\begin{figure}
\centering
  \includegraphics[width=1\linewidth]{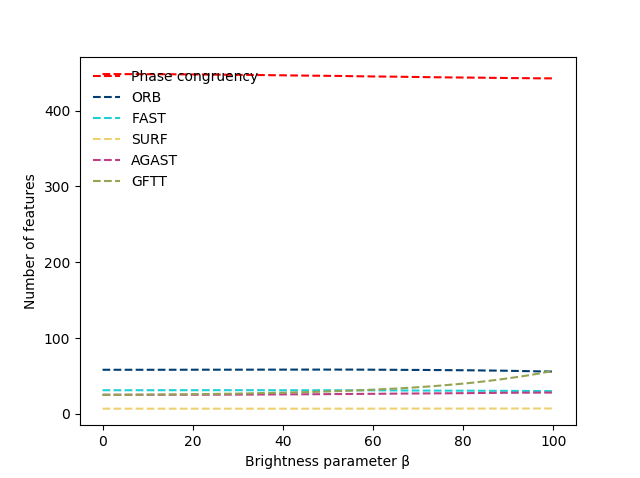}
  \caption{Average number of features detected for each image by some features extractors}
  \label{fig:numbers}
\end{figure}

So we took 1000 images (80x60) acquired using lepton 2, then we simulated brightness changes by varying the parameter $\beta$ from 0 to 100. For each image, we computed the number of features detected for each $\beta$. Let $\Omega_{\beta}$ be the number of matches for a value of brightness control $\beta$. The Figure \ref{fig:numbers} represents the variation of the $\Omega_{\beta}$ according to $\beta$.

We also needed to evaluate the features extraction method in terms of re-detection.

For this, first, we computed the features at $\beta$=0. Then by increasing $\beta$, we estimated this robustness by counting the percentage of features still detected at the same location as for $\beta$=0. We called this percentage features a re-detection rate. Figure 6 shows the Features re-detection rate vs. brightness change for one image of the Tvvlgo dataset \cite{VideoTraz}.

\begin{figure}
\centering
  \includegraphics[width=1\linewidth]{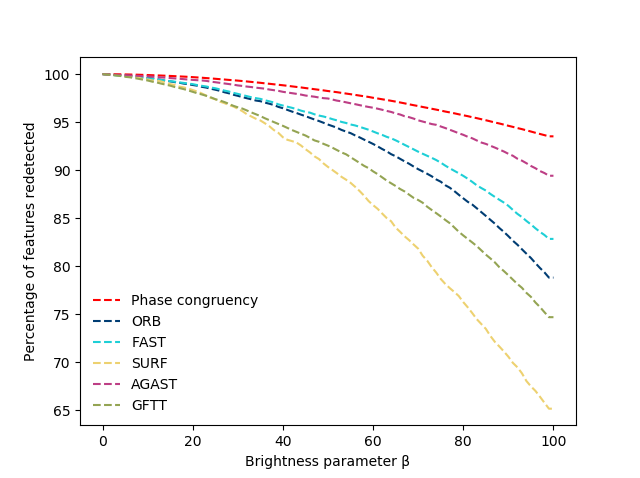}
  \caption{Features re-detection rate}
  \label{fig:redetection}
\end{figure}

We compared PhaseCong(0.1) with other feature extractors such as ORB, FAST, SURF, AGAST, and GFTT. Our results also confirmed those of Hajebi et al~\cite{Hajebi2007} assessing that the features detected by phase congruency from thermal images present the advantage to be more stable than the others to illumination changes (Fig. \ref{fig:numbers} and \ref{fig:redetection}). Figure \ref{fig:numbers} shows that given different values of illumination, phase congruency is the feature extraction method that can extract the highest number of features. Figure \ref{fig:redetection} shows that phase congruency is the method with the best re-detection rate. While detecting more features, phase congruency is also more robust to brightness changes than the other classical methods.

\subsection{Stereo matching}
\label{ssec:ExpStereoMatching}

To evaluate our stereo matching method, we have performed a stereo matching on 15 low-resolution $80 \times 60$ stereo pairs provided by our cameras (Tvvglo \cite{VideoTraz}). The stereo matching method explained in the section~\ref{sec:SubPixelMatchingFramework} was applied. To verify the accuracy of the method, we manually counted the number of mismatches given the number of matches. For each feature extracted in $I_l$, we visually verified if the matched feature in $I_r$ was inside a $5 \times 5$ window centered on the estimated disparity. The results showed that we had a percentage of mismatches of less than $1 \%$ through all of the 15 image pairs.

\begin{figure*}
\centering
 \includegraphics[width=\linewidth]{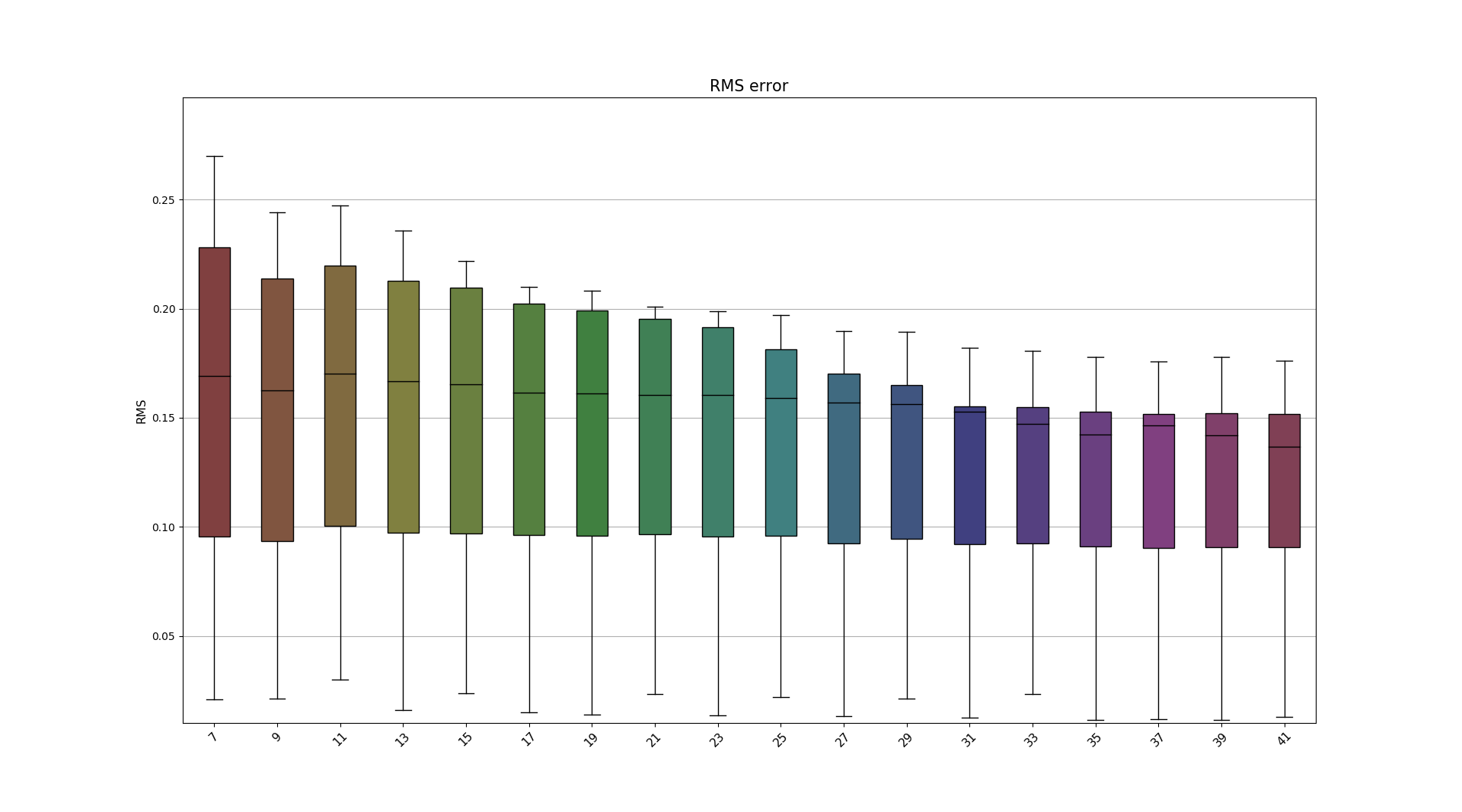}
\caption{Box plots of the root-mean-square deviation (RMSD) of the matching error (in pixel) vs. sub-images window size}
\label{fig:box_plot_rms}
\end{figure*}

\begin{figure*}
     \centering
     \begin{subfigure}[b]{0.475\textwidth}
        \centering
        \includegraphics[width=\textwidth]{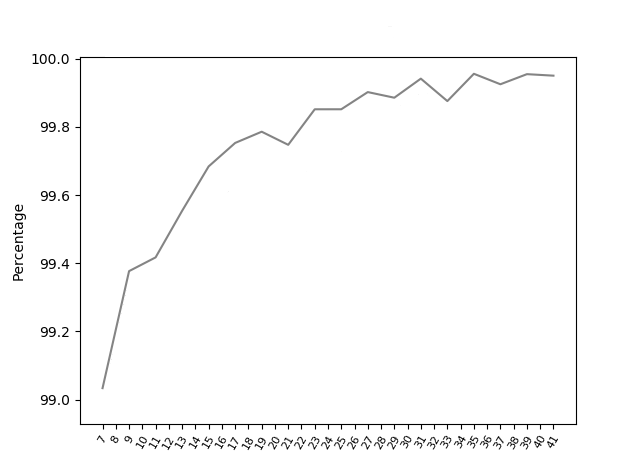}
        \caption[Network2]%
        {{\small Percentage of matching with an error less than 0.5 pixels}}   
        \label{fig:first_05_512}
     \end{subfigure}
     \hfill
     \begin{subfigure}[b]{0.475\textwidth} 
        \centering 
        \includegraphics[width=\textwidth]{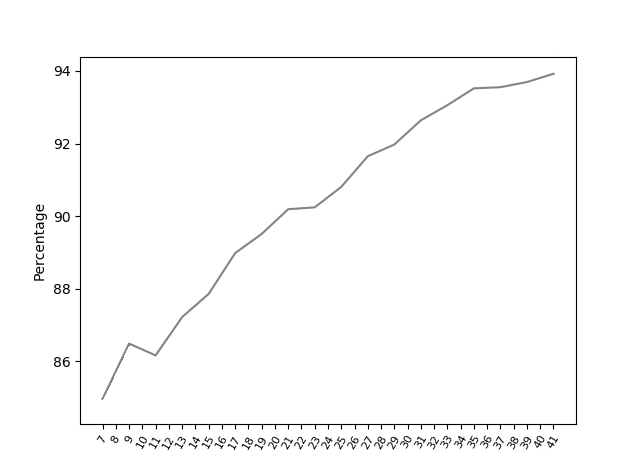}
        \caption[]%
        {{\small Percentage of matching with an error less than 0.25 pixels}}   
        \label{fig:first_025_512}
     \end{subfigure}
     \vskip\baselineskip
     \begin{subfigure}[b]{0.475\textwidth}  
        \centering 
        \includegraphics[width=\textwidth]{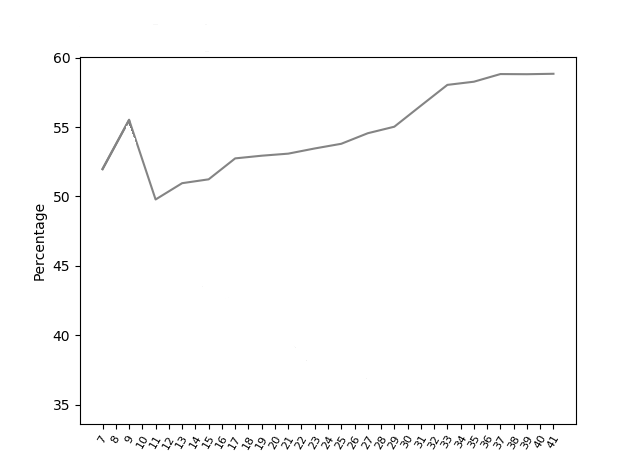}
        \caption[]%
        {{\small Percentage of matching with an error less than 0.1 pixels}}   
        \label{fig:first_01_512}
     \end{subfigure}
     \quad
     \begin{subfigure}[b]{0.475\textwidth}  
        \centering 
        \includegraphics[width=\textwidth]{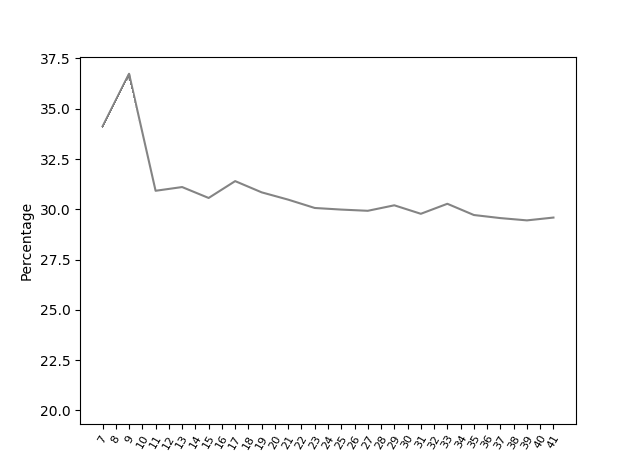}
        \caption[]%
        {{\small Percentage of matching with an error less than 0.05 pixels}}   
        \label{fig:first_005_512}
     \end{subfigure}
     \caption[ The average and standard deviation of critical parameters ]
     {\small  Impact of the sub-images window size (abscissa) on the rate of good matching (ordinate in \%) respects to a specific level of precision} 
     \label{fig:first_percentage_512}
   \end{figure*}

   \begin{figure*}
     \begin{subfigure}[b]{0.475\textwidth}
        \centering
        \includegraphics[width=\textwidth]{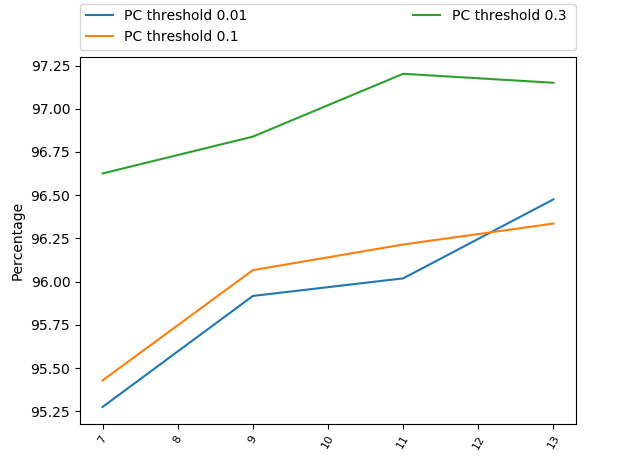}
        \caption[Network2]%
        {{\small Percentage of matching with an error less than 0.5 pixels}}   
        \label{fig:first_05_80}
     \end{subfigure}
     \hfill
     \begin{subfigure}[b]{0.475\textwidth} 
        \centering 
        \includegraphics[width=\textwidth]{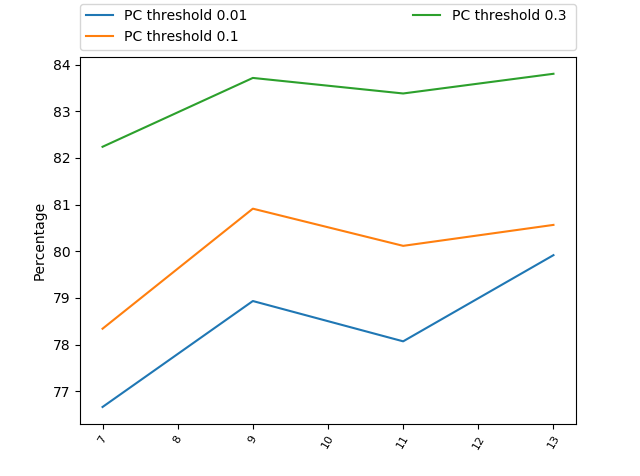}
        \caption[]%
        {{\small Percentage of matching with an error less than 0.25 pixels}}   
        \label{fig:first_025_80}
     \end{subfigure}
     \vskip\baselineskip
     \begin{subfigure}[b]{0.475\textwidth}  
        \centering 
        \includegraphics[width=\textwidth]{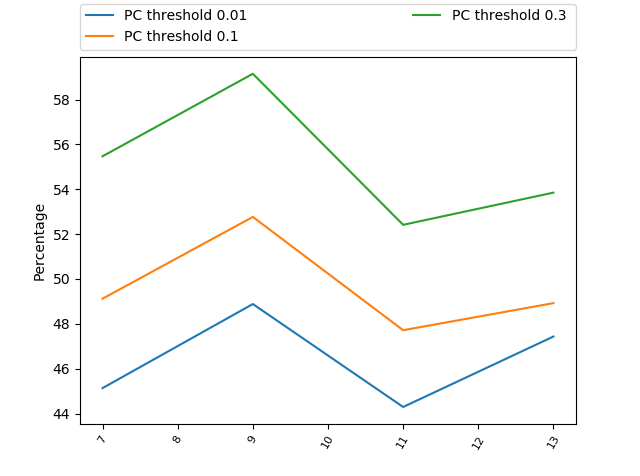}
        \caption[]%
        {{\small Percentage of matching with an error less than 0.1 pixels}}   
        \label{fig:first_01_80}
     \end{subfigure}
     \quad
     \begin{subfigure}[b]{0.475\textwidth}  
        \centering 
        \includegraphics[width=\textwidth]{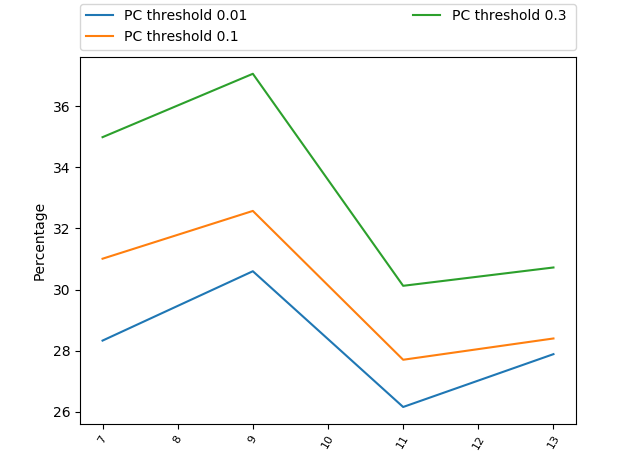}
        \caption[]%
        {{\small Percentage of matching with an error less than 0.05 pixels}}   
        \label{fig:first_005_80}
     \end{subfigure}
     \caption[ The average and standard deviation of critical parameters ]
     {\small Percentage of correct matches (ordinate in \%) with a given level precision $\tau$ according to the sub-images window size (abscissa) and the phase congruency moments threshold (PC threshold)} 
     \label{fig:flir_percentage_80}
   \end{figure*}

\subsection{Sub-pixel matching}
\label{sec:ExpSubPixelMatching}

As mentioned in section~\ref{ssec:SubPixelMatchingMeth}, one of the key parameters is the size $W$ of the sub-images windows in which the phase correlation is computed. In the state of the art, the sizes of sub-images vary a lot. For images with good resolution such as $4000 \times 4000$, a $W$ of $41 \times 41$ can be chosen~\cite{Ma2017}. But these window sizes must not only be set regarding the size of the images but also depends on the type of the scene and how the disparity varies through the whole image. Nevertheless, it would be unreasonable to use a $W$ of $41 \times 41$ when the image size is $80 \times 60$. 

In fact, we set 2 experiments, one on good resolution images in order to estimate the sole impact of $W$ on the accuracy of sub-pixel matching and one on our low resolution images to prove the validity of our method according to the end application and to estimate the combined impact of the phase congruency threshold $\gamma$ and $W$ on the sub-pixel matching accuracy. We designed the following experiment to estimate the sub-pixel matching accuracy: for a specific thermal image $I_1$, we created a second image $I_2$ by shifting the information of $I_1$ by $\Delta$ pixels along the $x$-axis. The accuracy can be determined according to the distance between an estimated sub-pixel disparity to $\Delta$ (the true disparity). We called it \emph{matching error}. We also created a \emph{matching precision rate} measurement. For this, we set a precision threshold of $\tau$. Given an estimated disparity $\delta$ if $\left | \delta - \Delta \right | \leq \tau $ the match is considered as correct else as a mismatch. The \emph{matching precision rate} is the ratio between the number of correct matches (relative to a level of precision $\tau$) to the total number of features.

 For the good resolution thermal images, we used a $512 \times 512$ image of people coming from~\cite{Wu2014}. We shifted these images by $\Delta \in \left \{ 20,20.125,20.250,...,25 \right \}$ pixels. The matching errors were computed on all the features for all the set of $\Delta$. Fig.~\ref{fig:box_plot_rms} shows the box plot of the root-mean-square deviation (RMSD) of the matching errors as a function of the window size $W$, with $W \in \left\{ 7, 9, \ldots, 41 \right\}$. The impact of the window size on the rate of good matching respects to a specific level of precision $\tau$ can be seen in Fig.~\ref{fig:first_percentage_512}. We drew this curves for 4 levels of precision: $\tau = \left\{0.05 , 0.01, 0.025 , 0.5 \right\}$ pixels.

The low resolution ($80 \times 60$ pixels) images were acquired with our FLIR Lepton 2 camera system (\cite{VideoTraz}). For each image we shifted them by $\Delta \in \left \{ 0,0.125,..,30 \right \}$pixels  with a stride of $0.125$ pixels. The size of our sub-images was in the range $W \in \left \{ 7 \times 7 ; 9 \times 9 ; 11 \times 11 \right \}$. The Figure \ref{fig:flir_percentage_80} shows the impact of the window size $W$ and the phase congruency moments threshold $\gamma$ on the rate of good matching respects to a specific level of precision $\tau$, $\tau \in\left \{ 0.5 , 0.25 , 0.1 , 0.05 \right \}$ pixels in our case. The value of the phase congruency moments threshold $\gamma$ is directly related to the number of features extracted on the images (Table~\ref{tab:compFeatures}). However, we had to find a good tradeoff between the number and significance of the extracted features.

We studied the impact of the sub-images window size in which the phase correlation is estimated on the accuracy of the sub-pixel localization. As shown in Fig.~\ref{fig:box_plot_rms}, using a wide sub-images window allowed us to get better precision. From $7 \times 7$ to $29 \times 29$, the precision was increased progressively, but from $31 \times 31, $ the gain in precision was no more very noticeable. The phase congruency magnitude represents the filtered information of the image, so using a wider window does not help necessary to get better precision. This result is also confirmed in Fig.~\ref{fig:first_percentage_512} and \ref{fig:flir_percentage_80} on the cases were low precision was sufficient. These Figures showed the percentage of correct matches at a certain precision using different sizes of window size. For low precision (errors below $\tau = 0.5$ or $\tau = 0.25$ pixels), the matching rate was increased with higher sub-images window size. This behavior was shared on both high and low-resolution images. The only difference between these 2 cases is that, for the matching of low-resolution images, we had to limit the sub-images window size to $13 \times 13$, which is already large compared to the $80 \times 60$ image size.

More surprisingly, by analyzing the three Figures accurately, it can be noticed that a window $9 \times 9$ offered better precision than an $11 \times 11$ one. This is even more visible on the high precision matching rates (errors below $\tau = 0.1$ or $\tau = 0.05$ pixels) curves (Fig.~\ref{fig:first_percentage_512}-c and d and \ref{fig:flir_percentage_80}-c and d). One explanation could be that the precision of the match did not only depend on the size of the window size but overall by the ratio between noise and relevant information in this window. In our framework, they are 2 processes that handle the image noise: 1) the phase congruency magnitude represents filtered information of the image and 2) a low pass filter the phase correlation. In this latter case, the filter bandwidth was directly set proportional to the sub-images window size ($U = 0.5 W$ in (\ref{eq:normalizedCrossPowerSpectrum})). It seemed like that for $W = 9 \times 9$, the ratio between the useful information (on low spatial frequency in the cross-power spectrum (\ref{eq:normalizedCrossPowerSpectrum})) and noise (on the high spatial frequency) reached a local maxima. If this hypothesis is confirmed, more noise is included in the phase correlation for higher $W$, degrading so its shape.

The expected precision also had a direct impact on the matching rate. We can see in Fig.~\ref{fig:first_percentage_512} that for the high-resolution image, almost all the features (more than 99\%) are matched with a precision lower than 0.5 pixels regardless of the window size. This matching rate decreased to around $90\%$, $55\%$, and $33\%$ for respectively a precision lower than 0.25, 0.1, and 0.05 pixels. We can notice the same behavior for the low-resolution image with matching rates (in the case of the best phase congruency threshold) around $97\%$, $83\%$, $55\%$ and $34\%$ for respectively a precision lower than 0.5, 0.25, 0.1 and 0.05 pixels. Moreover, we noticed the same matching rate ($\approx 33\%$), whatever the resolution of the images when a high matching precision is expected.

The choice of the phase congruency moments threshold also had an impact on the matching rates. Because of the low resolution of our images, we wanted to increase the number of features using the lower phase congruency moments threshold. This was the case, as shown in Table~\ref{tab:compFeatures}. The number of extracted features increased from 275 to 1793 when we decreased the threshold of $\gamma$ from 0.3 (as recommended in~\cite{kovesi2003}) to 0.01. Unfortunately, the matching rates for a specific precision, also decreased when we decreased the threshold, as we can see in Fig.~\ref{fig:flir_percentage_80}. 
Decreasing the threshold brought less stable features to match. Depending on the application, a good tradeoff has to find between the number of matched features and their reliability.

To summarize, in our specific low resolution stereo thermal camera case, a sub-images window size window of $9 \times 9$ can be a good tradeoff between the accuracy and the phase correlation computation time. With this configuration, we were able to match around $97\%$ of the features with a precision of at least 0.5 pixels. If a higher sub-pixel matching accuracy is needed, this rate falls to $55\%$ or $34\%$ for a respectively a precision less than 0.1 or 0.05 pixels. In this case, an external outlier rejection process based on the image or the 3D scene content should be added during or after the matching~\cite{Wolff2016}. 
\subsection{3D reconstruction}


\subsubsection{Importance of sub-pixel matching}

After matching, an error on the estimation of the disparity can have high consequences for the stereo reconstruction, especially to estimate the distance of a 3D point to the cameras (Z direction) \cite{haller2011design} accurately. In this section, we wanted to estimate the range of error along the Z direction when we estimated the disparity with an error of one pixel (in pixel resolution)  and also the evolution of this range using sub-pixel disparity estimation (error of 0.5 or 0.1 pixels).
For this, we chose a pair of matched points, which gave a distance Z=3350 mm after stereo reconstruction for a disparity d. If we have an error of 1 pixel (a disparity of $d_{new} = d\pm1$ pixel), the estimation of Z was spread from  Z = 3108 mm to 3590 mm (a range of 482 mm). Being wrong in a range of 482 mm in a human body reconstruction can be severe for many surveillance applications. For an error of $d_{new} = d\pm0.5$ the range was reduced to 270 mm and for $d_{new} = d\pm0.1$ pixel to 52 mm. 
Such results explain why sub-pixel matching is essential in our low-resolution context.

\subsubsection{Evaluation of the whole framework}

\begin{figure*}
     \begin{subfigure}[b]{0.31\textwidth}
        \centering
        \includegraphics[width=\textwidth]{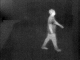}
        \caption[Network2]%
        {{\small Left image of the 507 th pair of Tvvlgo}}   
        \label{fig:left507}
     \end{subfigure}
     \hfill
     \begin{subfigure}[b]{0.31\textwidth} 
        \centering 
        \includegraphics[width=\textwidth]{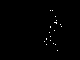}
        \caption[]%
        {{\small Matches re-projected using ORB (\#24)}}   
        \label{fig:orb507}
     \end{subfigure}
     \hfill
     \begin{subfigure}[b]{0.31\textwidth} 
        \centering 
        \includegraphics[width=\textwidth]{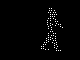}
        \caption[]%
        {{\small Matches re-projected using ST (\#80)}}   
        \label{fig:pc507}
     \end{subfigure}

     \vskip\baselineskip
     
     \begin{subfigure}[b]{0.31\textwidth}
        \centering
        \includegraphics[width=\textwidth]{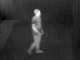}
        \caption[Network2]%
        {{\small Left image of the 547 th pair of Tvvlgo}}   
        \label{fig:left547}
     \end{subfigure}
     \hfill
     \begin{subfigure}[b]{0.31\textwidth} 
        \centering 
        \includegraphics[width=\textwidth]{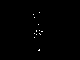}
        \caption[]%
        {{\small Matches re-projected using ORB (\#18)}}   
        \label{fig:orb547}
     \end{subfigure}
     \hfill
     \begin{subfigure}[b]{0.31\textwidth} 
        \centering 
        \includegraphics[width=\textwidth]{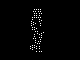}
        \caption[]%
        {{\small Matches re-projected using ST (\#71)}}   
        \label{fig:pc547}
     \end{subfigure}

     \caption{3D points projected in the images space with Z as the value of the pixel.}
     \label{fig:3dpoints}
   \end{figure*}

\begin{figure}
\centering
  \includegraphics[width=\linewidth]{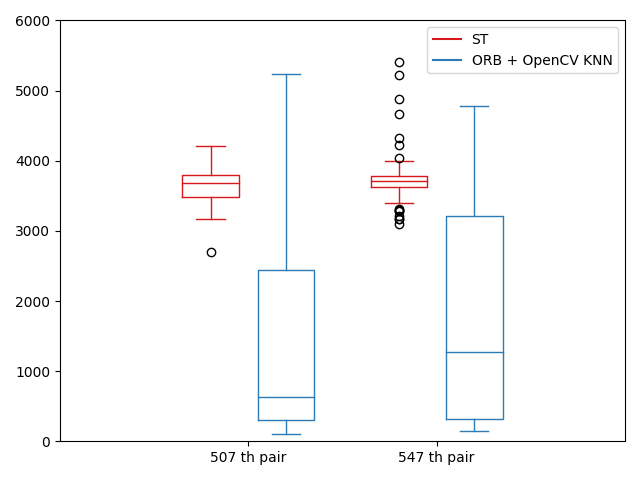}
  \caption{Box plot in the value of Z for our method ST and ORB + OpenCV KNN.}
  \label{fig:boxcompared}
\end{figure}

\begin{table}
\centering
\begin{tabular}{|l|l|l|}
\hline
            & \begin{tabular}[c]{@{}l@{}}ORB + OpenCV\\  KNN (Baseline)\end{tabular} & ST (Our method) \\ \hline
507 th pair & 1523.94 $\pm$  1612.76                                                                & 3648.10 $\pm$ 256.43      \\ \hline
547 th pair & 1800.85  $\pm$ 1538.92                                                               & 3746.56 $\pm$ 391.44         \\ \hline
\end{tabular}
\caption{Mean value and standard deviation of Z after triangulation. Values are in millimeters.}
\label{tab:res1}
\end{table}

In this section, we tried to compare the whole framework of our method ST versus a method available in classical computer vision libraries. As a feature extractor, we chose ORB because this feature extractor has already been proved to be more robust than other \cite{karami2017image}. For the matching, we used KNN because of the sparsity of the features and also because it estimates the disparity in sub-pixel precision. We implemented this framework (we called \textbf{ORB + KNN matching}) using the functions available in OpenCV. For ST, we set $\gamma = 0.1$ and for phase correlation we used a sub-image size of $9 \times 9$.

We applied these methods on two image pairs (Fig.~\ref{fig:left507} and Fig.~\ref{fig:left547}) we selected from the Tvvlgo dataset. In these images, the person was approximately at a distance of 3-4 meters. We compared the results of these methods according to two criteria: the numbers of matched features and the consistency of the 3D reconstruction, especially along the Z direction.

The Figure \ref{fig:3dpoints} represents the matches projected on the left images  when using ORB + KNN (\ref{fig:orb507} and \ref{fig:orb547}) and ST (\ref{fig:pc507} and \ref{fig:pc547}). First of all, we can see that using ST, we have 4 $times$ more matches than ORB + KNN: 80 vs. 24 and 71 vs. 18 respectively for the 507 th pair and the 547 th pair. It can also be noticed that the matches are better distributed over the whole human shape than using ST than using ORB + KNN. This is due to the fact that phase congruency can extract more features from thermal images than ORB, as reported in Table \ref{tab:compFeatures}.

For all the matches, we performed triangulation to estimate the 3D position of the point. We made statistics on the estimated depth Z. Figure \ref{fig:boxcompared} shows the boxplot of the distributions of Z for both methods. The mean and the standard deviation of these distributions are given in Table \ref{tab:res1}. The results obtained using ORB + KNN seems to be inconsistent (very wide distribution of Z in a range of about 5000 mm).
On the contrary, for ST, the distribution is more compact. The median and mean values are around 3500 mm, which is consistent with the experimental conditions, and the standard deviations (256 mm and 391 mm) are more reliable concerning human body proportion. The results output by ORB + KNN can be explained by the fact that the matches are not correct, leading to very bad triangulation outputs.

\section{Conclusion} 
\label{sec:Conclusion}

In this paper,  we proposed our method ST which is a sub-pixel stereo matching method adapted to thermal images. Thermal images have the disadvantages of being less textured compared to gray-scale or color images. Besides this lack of texture, low-cost thermal cameras can have a very low resolution ($80 \times 60$ pixels in our case), which is detrimental to an accurate stereo reconstruction. To overcome these limitations, we proposed a framework composed of a robust feature extraction method based on phase congruency, a robust rough matching process based on Lades distance between the extracted features, and a refined sub-pixel matching process based on phase correlation. When applied to low-resolution thermal images, our feature extraction method was able to extract more features than state of the art methods. As well, our sub-pixel matching method was able to match around 97\% of extracted features with an (average) error under 0.5 pixels. For about 55\% of the features, the matching error was even below 0.1 pixels. Such a level of accuracy is necessary for the stereo reconstruction of a 3D scene or the 3D localization of objects. With our stereo setup, a precision bellow 0.1 pixels corresponds to a maximal error of $\approx 51 $ mm in the depth direction. With our framework, such a level of precision seems now achievable even for very low-resolution thermal stereo cameras. We also compared ST versus a classical method in state of the art (ORB + KNN) for 3D reconstruction. Our method showed more consistent results than ORB + KNN.

Once the sub-pixel matching is performed, it could be essential to compare our method in 3D localization using ground truth values. Such evaluation will show us how sub-pixel matching is improving 3D vision.

\section*{Acknowledgments}    
 
This work was part of the PRuDENCE project (ANR-16-CE19-0015) which has been supported by the French National Research Agency (ANR).


\bibliographystyle{cas-model2-names}

\bibliography{references}